\documentclass{bmvc2k}

\title{Video Region Annotation with Sparse Bounding Boxes} 

\addauthor{Yuzheng Xu}{yxu@vision.ist.i.kyoto-u.ac.jp}{1}
\addauthor{Yang Wu}{wu.yang.8c@kyoto-u.ac.jp}{1}
\addauthor{Nur Sabrina binti Zuraimi}{sabrina@vision.ist.i.kyoto-u.ac.jp}{1}
\addauthor{Shohei Nobuhara}{nob@i.kyoto-u.ac.jp}{1}
\addauthor{Ko Nishino}{kon@i.kyoto-u.ac.jp}{1}

\addinstitution{
 Kyoto University\\
 Kyoto, Japan
}

\runninghead{Xu et al.}{Video Region Annotation with Sparse Bounding Boxes}

\usepackage{graphicx}
\usepackage{comment}
\usepackage{amsmath,amssymb} 
\usepackage{multirow}                
\usepackage{booktabs}
\usepackage{todonotes}
\usepackage{enumitem}

\usepackage{xcolor}
\usepackage{here}

\begin{document}

\maketitle

\begin{figure}[!ht]
\begin{center}
  \includegraphics[width=1.0\linewidth]{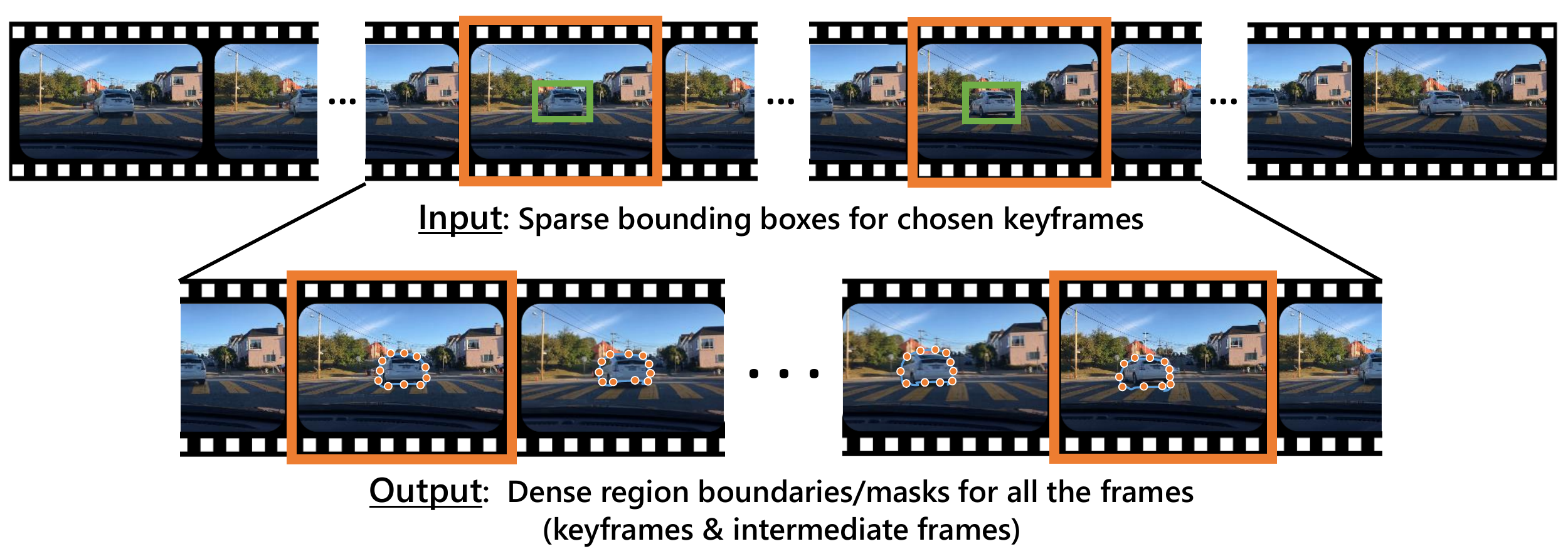}
\end{center}
\caption{Our goal is to derive a video region annotation tool that can automatically annotate dense per-frame region boundaries from sparse user-provided bounding boxes given for sparse keyframes.}
\label{fig:teaser}
\end{figure}

\begin{abstract}

Video analysis has been moving towards more detailed interpretation (e.g. segmentation) with encouraging progresses. These tasks, however, increasingly rely on densely annotated training data both in space and time. Since such annotation is labour-intensive, few densely annotated video data with detailed region boundaries exist. This work aims to resolve this dilemma by learning to automatically generate region boundaries for all frames of a video from sparsely annotated bounding boxes of target regions. We achieve this with a Volumetric Graph Convolutional Network (VGCN), which learns to iteratively find keypoints on the region boundaries using the spatio-temporal volume of surrounding appearance and motion. The global optimization of VGCN makes it significantly stronger and generalize better than existing solutions. Experimental results using two latest datasets (one real and one synthetic), including ablation studies, demonstrate the effectiveness and superiority of our method.

\end{abstract}

\section{Introduction}
\label{sec:intro}

Advances in deep learning techniques have brought about remarkable progress in many computer vision tasks such as detection, segmentation, tracking, and recognition. One major caveat with most deep learning algorithms is that they need to be trained with a huge amount of data that have been carefully labeled with ground truth~\cite{Everingham:2015:PVO:2725268.2725369,DBLP:journals/corr/LinMBHPRDZ14,DBLP:journals/corr/ZhouZPFBT16}. 
Furthermore, in many applications such as autonomous driving, visual analysis has to be done on every captured frame for real-time processing or for tasks that require dense spatio-temporal information. For these, dense per-frame region-level annotation becomes essential for training the models.

Manually annotating detailed region boundaries for every video frame is a highly time-consuming, tedious, if not impossible, task. To our best knowledge, no publicly available dataset offers per-frame annotation. The lack of densely annotated video data has limited the research on detailed region-level video analysis and have forced researchers to explore image-based models instead. Frame-wise processing, however, misses the spatial-temporal relationships and can lead to inferior results. As such, dense per-frame region annotation with an affordable and efficient means becomes critical. Bounding box is a widely used and rather cheap supervision. What if we only need annotators to provide region bounding boxes for sparsely chosen keyframes and then the annotation tool automatically generates boundaries for the region of interest in every frame, as illustrated in Figure \ref{fig:teaser}?

We introduce a novel dense video annotation method that only requires sparse bounding-box supervision. We fit an iteratively deforming volumetric graph to the video sub-sequence bounded by two chosen keyframes, so that its uniformly initialized graph nodes gradually move to the key points on the sequence of region boundaries. The model consists of a set of deep neural networks, including normal convolutional networks for frame-wise feature map extraction and a volumetric graph convolutional network for iterative boundary point finding. By propagating and integrating node-associated information (sampled from feature maps) over graph edges, a content-agnostic prediction model is learned for estimating graph node location shifts. The effectiveness and superiority of the proposed model and its major components are demonstrated on two latest public datasets: a large synthetic dataset Synthia and a real dataset named KITTI-MOTS capturing natural driving scenes.

\section{Related Work}



\paragraph{Region Annotation vs. Segmentation}
Since one can easily get confused by the relationship between our work and large amounts of existing works on segmentation tasks (including semantic segmentation \cite{J.Li2019,Ding2019,Paul2019,Seyed2018}, object segmentation \cite{jain2017fusionseg}, and instance segmentation \cite{VIS2019}), we first clarify the difference.
Annotation is the process of labeling data to be used for machine learning algorithms, including training and evaluation. Whilst there have been many studies in learning from unlabeled data \cite{Cho_2015_CVPR,Wang_2015_ICCV,Lee_2017_ICCV}, many state-of-the-art algorithms still need some sort of labeled data for training, and in any case quantitative performance evaluation generally requires ground-truth labels. On the other hand, segmentation is the process of predicting pixel-level class labels. 
The main difference between annotation and segmentation is that \emph{annotation is for building a dataset whilst segmentation is a vision task model trained on an annotated dataset. As such, any segmentation method would benefit from a better annotated data and hence the annotation tool}. This work focuses on region annotation for videos, aiming to alleviate the burden of the annotator to help make the process of creating ground truth data easier, and thus support the development of new video analysis models, including those for image/video based segmentation.

\paragraph{Single-Image Annotation Tools}
In general, one can still choose to annotate each frame using image annotation tools. Representative works are briefly discussed here. One of the earliest annotation tool that aimed to cut down the time required to annotate was GrabCut \cite{GrabCut} which does interactive foreground/background segmentation in still images using bounding boxes and foreground and background marking strokes as its inputs. Polygon-RNN \cite{PolyRNN} and Polygon-RNN++ \cite{acuna2018efficient} use a CNN-RNN architecture to sequentially trace object boundaries given a bounding box. The RNN can only output one vertex at a time which could mean slow inference time depending on the number of vertices to be inferred. In subsequent work, Curve-GCN \cite{Curve-GCN-CVPR19} attempts to get around this limitation by modeling object annotation as a boundary control point regression problem and using graph convolutions to do the joint regression for all the control points (i.e., graph nodes). It was demonstrated to be faster and also more effective than Polygon-RNN. Since these models/tools do not use temporal information among successive frames, simply extending them for the desired video region annotation task is likely to be inferior than our proposed solution. To demonstrate it, we build two extensions of the state-of-the-art model Curve-GCN \cite{Curve-GCN-CVPR19} and compare them with our proposed model in the experiment section (section \ref{sec:experiment}).

\paragraph{Video Annotation Tools} Annotating objects in video is not as straight-forward as annotating them in images as it requires observing their motion paths and taking into account the possibility of change in shape over time. One of the earliest video annotation tools publicly available is VATIC \cite{10.1007/978-3-642-15561-1_44} which uses inter-frame interpolation to generate bounding boxes automatically. Bounding boxes, however, are not enough for detailed analyses including pixel-wise segmentation. There have also been efforts on annotating regions in videos using active contours \cite{TouchCut}, approximation of closed boundaries using polygons \cite{Bianco:2015:ITM:2735612.2741539}
and partition trees \cite{Giro-i-Nieto2010}. Despite the differences in getting the region boundaries within a frame, all these tools do some kind of annotation propagation or interpolation across video frames to achieve video annotation. Instead, our proposed model jointly optimizes the boundaries in all frames.







\section{Volumetric Graph Convolutional Network (VGCN)}


\begin{figure}[t]
\begin{center}
  \includegraphics[width=1.0\linewidth]{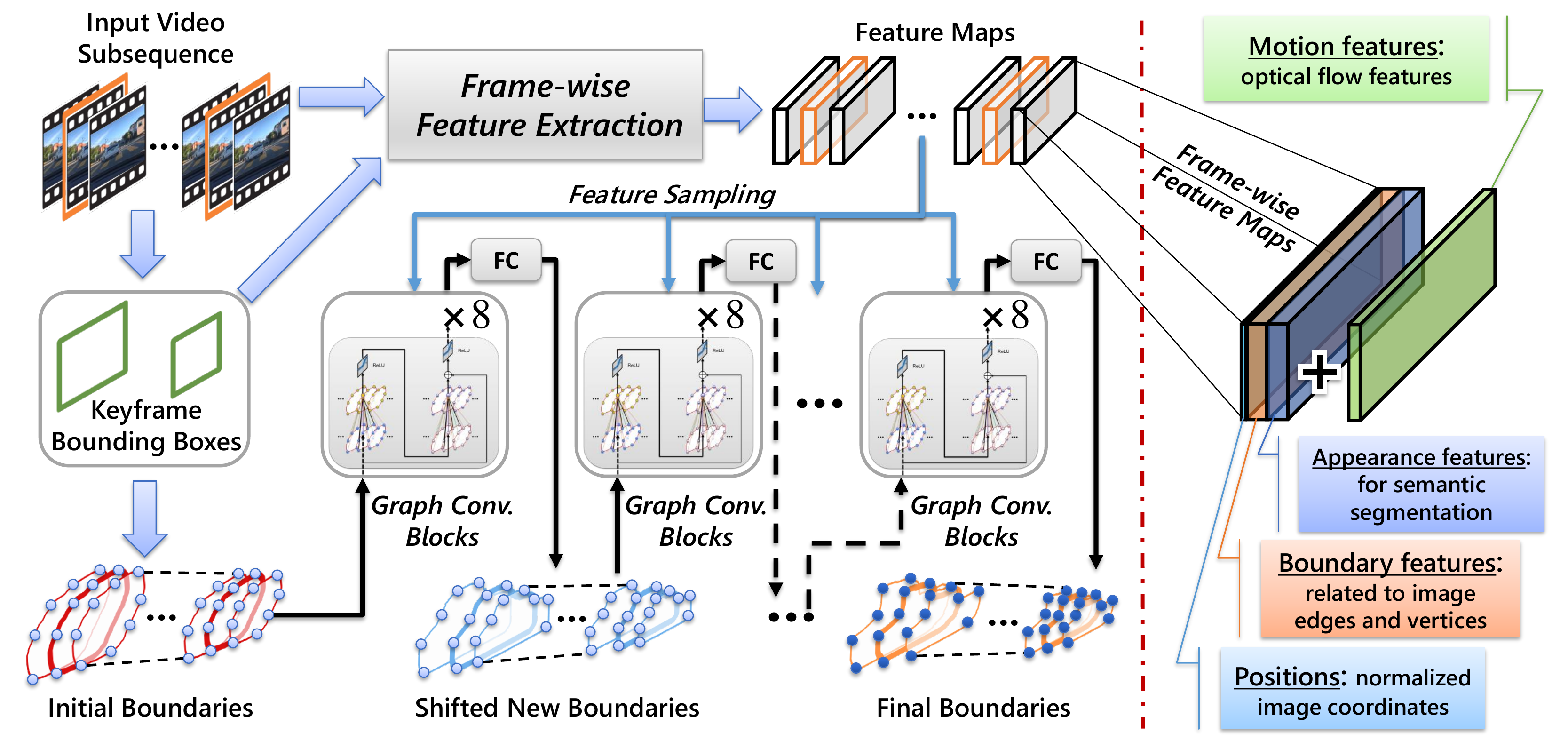}
\end{center}
\caption{The overall framework of Volumetric Graph Convolutional Network (VGCN). Per-frame feature maps are illustrated on the right.}
\label{fig:framework}
\end{figure}

Our aim is to automatically generate dense per-frame region boundary labels for all the regions of interest in all video frames, assuming only the bounding boxes of the target regions in sparse keyframes are given by human annotator(s). The whole task can be decomposed into subtasks each of which focuses on a key step: generating dense per-frame region boundary labels for a sub-sequence of video frames bounded by two keyframes, where a single region inside the target appears across all the frames, as shown in Figure \ref{fig:teaser}. This is a reasonable setting as human annotators can scan the whole video before annotation and place keyframes to cut the whole sequence of the target region (object, object part, or even stuff) into sub-sequences with reasonably consistent region shapes. 

We have three desiderata for a video annotation model. First, besides the raw video data, the model can only assume \emph{sparse bounding boxes} as the input during testing and inference. Second, the method should be applicable to \emph{arbitrary regions} (of different shapes and contents) appearing in sub-sequences of arbitrary lengths. Last, \emph{spatial-temporal inference} should be employed to ensure global joint optimization. In this paper, we propose a novel model which we refer to as Volumetric Graph Convolutional Network (VGCN) that meets all three requirements. It takes as input a pair of bounding boxes from two keyframes for data cropping and normalization, as well as initialization. Its local graph connections (i.e., edges) and weight sharing over same types of connections allow uniform formulation and robust modeling of arbitrary local shapes. The volumetric graph convolutions integrate and propagate information spatially and temporally, leading to global spatial-temporal inference and the ability to handle arbitrary video of various lengths.



\subsection{Overall Framework}


As shown in Figure \ref{fig:framework}, given an input video sub-sequence bounded by two chosen keyframes, the annotator-provided keyframe bounding boxes are used to crop the video frames, normalize them (following \cite{Curve-GCN-CVPR19}), and extract frame-wise feature maps whose contents are shown on the right. The bounding boxes also help VGCN initialize the locations of its volumetric nodes that correspond to the keypoints of desired boundaries of all video frames. Then the model samples features from the feature maps according to the node (i.e., boundary keypoint) locations, and such sampled features are fed into a group of graph convolutional blocks (8 of them in our implementation) for information integration and propagation. A fully-connected (FC) layer is adopted to map the updated features of each node to its predicted location shifts. After the actual shifting of node locations, another round of feature sampling and graph convolutions can be applied to predict a new round of location shifts. This process can be iterated several times to ensure an accurate fit to the actual region boundaries.

\subsection{Graph Structure}

\begin{figure}[t]
\begin{center}
  \includegraphics[width=1.0\linewidth]{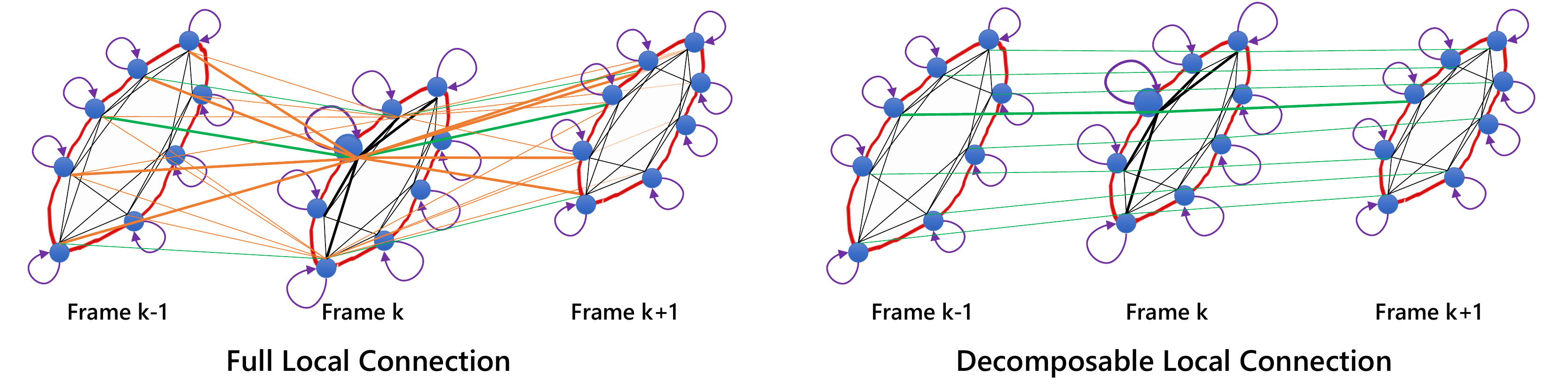}
\end{center}
\caption{Graph structures of two variants of VGCN, full local connection (recommended) and decomposable local connection (simplest), illustrated with three adjacent frames. Note that all the nodes of an intermediate frame will have exactly the same number of edges. Some edges for ``full local connection'' are omitted for better visibility.} 
\label{fig:graph_structure}
\end{figure}

Suppose the sub-sequence where a target region exists is bounded by two keyframes with a sparsity factor $K$ indicating the frame ID difference between them. The task is to find the region boundary in each frame. The 1st frame and the ``$K+1$''-th frame are the keyframes with bounding-box supervision. Assume the shape of the region boundary in each frame can be well-represented by $N$ control points $\mathcal{V}_k = \{\mathbf{cp}_k^0, \dots, \mathbf{cp}_k^{N-1}\}$, where $k$ indicates the $k$-th frame and $\mathbf{cp}_k^i = [x_k^i, y_k^i]^T$ is the location of the $i$-th control point in this frame, we construct a volumetric graph $G_v = (\mathcal{V}, \mathcal{E})$ covering all the frames of the sub-sequence, for which a three-frame slice is illustrated in Figure \ref{fig:graph_structure}. Let $\mathcal{V} = \bigcup_{k=1}^{K+1} \mathcal{V}_{k}$ denote the graph nodes which are the union of control points from all the frames, we define the edge set $\mathcal{E} = \mathcal{E}_s \cup \mathcal{E}_t$ by introducing two types of connections for each node $\mathbf{cp}_k^i$. The spatial connections $\mathcal{E}_s$ cover both the node's self-connection and the links between the node and its four neighboring nodes (the black lines in Figure \ref{fig:graph_structure}), while the temporal connections $\mathcal{E}_t$ link the node to its corresponding nodes in the two neighboring frames (cross-frame green lines) and optionally also those nodes' four spatial neighbors (orange lines). Depending on the temporal links, the former case is called ``decomposable local connection'' as the spatial and temporal links are orthogonal, and the later is called ``full local connection'' or simply ``full-connection'' which is the recommended structure. These spatial-temporal connections enable effective and efficient information integration and propagation among the graph nodes. For a better model-data fit, we add one more frame to each end of the sub-sequence. As a result, our model deals with $K+3$ frames and has $\mathcal{V} = \bigcup_{k=0}^{K+2} \mathcal{V}_{k}$.

\subsection{Feature Extraction and Representation}


With the two keyframe bounding boxes, we get the bounding boxes for other frames with linear interpolation. Then a 15\% margin is added to each size of the bounding box to ensure a sufficient coverage of data even when the bounding boxes are tight. The extended bounding boxes are used to crop the frames and normalize the cropped areas to uniform sizes (with spatial coordinates normalized to [0,1] by [0,1]). Following \cite{Curve-GCN-CVPR19}, we extract three types of features from the normalized data as shown in Figure \ref{fig:framework}: appearance features computed using the DeepLab-v2 \cite{DeepLabv2} model pre-trained, and fine-tuned for semantic segmentation on ImageNet and PASCAL, respectively, and boundary features computed using additional branches to predict the probability of existence of an object edge/vertex on a 28 x 28 grid and position features (normalized pixel coordinates). In contrast to Curve-GCN \cite{Curve-GCN-CVPR19}, we add motion information into the feature extraction in two ways: concatenating the optical flow map (obtained by FlowNet2.0 \cite{FlowNet2}) to the original image data before feature extraction, and directly concatenating the flows to the three types of extracted features to form the final frame-wise feature maps. The first one contributes to the boundary features via early fusion, making it motion-aware, and the second one directly adds in the flow information via late fusion. In our implementation, we found that it is better to have the late fusion only involved in the first iteration of graph convolutions rather than all of them. This is likely due to the potentially noisy flows around the region boundaries. When the graph nodes get close to the boundaries in later iterations, such noise can obstruct the model from learning proper shifts.

\subsection{Graph Convolutional Block}

\begin{figure}[t]
\begin{center}
  \includegraphics[width=1.0\linewidth]{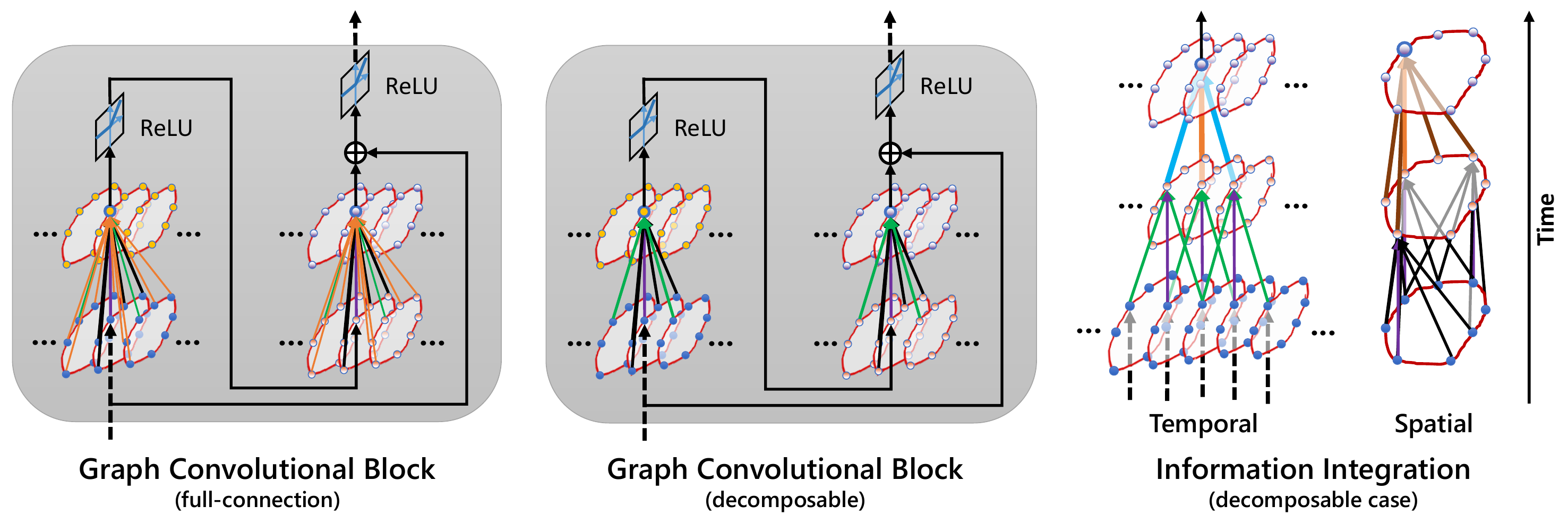}
\end{center}
\caption{The graph convolutional block of VGCN. Both full local connection (recommended) and decomposable local connection (simplest) are illustrated. The right subfigure shows how the information from graph nodes (boundary keypoints) is integrated spatially and temporally, using the decomposable model as an example.}
\label{fig:graph_conv_block}
\end{figure}

Following Curve-GCN, we also adopt the multi-layer GCN architecture, where each layer has a Graph ResNet structure as shown in Figure \ref{fig:graph_conv_block}. Mathematically, the graph convolution for an arbitrary node $\mathbf{cp}_k^i$ can be formulated as
\begin{equation}
    \hat{f}_{k,i}^l = W_o^l f_{k,i}^l+\sum_{\mathbf{cp}_k^j\epsilon \mathcal{E}_s(\mathbf{cp}_k^i) } W_s^l f_{k,j}^l + \sum_{\mathbf{cp}_{k-1}^j \epsilon \mathcal{E}_t(\mathbf{cp}_k^i) } {W_t^l f_{k-1,j}^l} + \sum_{\mathbf{cp}_{k+1}^j \epsilon \mathcal{E}_t(\mathbf{cp}_k^i) } {W_t^l f_{k+1,j}^l},
\end{equation}
where $W_o^l$, $W_s^l$, and $W_t^l$ are the weight matrices at layer $l$ to be learned for transforming the control point's own features, the features of its spatial neighbors (i.e., the nodes connected to $\mathbf{cp}_k^i$ via edges in $\mathcal{E}_s(\mathbf{cp}_k^i)$), and the features of its temporal neighbors (i.e., the nodes connected to $\mathbf{cp}_k^i$ via edges in $\mathcal{E}_t(\mathbf{cp}_k^i)$), respectively, and $\hat{f}_{k,i}^l$ is the updated feature for $\mathbf{cp}_k^i$ at layer $l$ after one round of information propagation. Note that in the ``full-connection'' case the two types of temporal edges (the ones connecting corresponding nodes, and the others) have different weight matrices.
After that, there is a ReLU activation $g_{k,i}^l = ReLU(\hat{f}_{k,i}^l)$.

Then, there is another round of convolution
\begin{equation*}
    \hat{g}_{k,i}^l = \tilde{W}_o^l g_{k,i}^l+\sum_{\mathbf{cp}_k^j\epsilon \mathcal{E}_s(\mathbf{cp}_k^i) } \tilde{W}_s^l g_{k,j}^l + \sum_{\mathbf{cp}_{k-1}^j \epsilon \mathcal{E}_t(\mathbf{cp}_k^i) } {\tilde{W}_t^l g_{k-1,j}^l} + \sum_{\mathbf{cp}_{k+1}^j \epsilon \mathcal{E}_t(\mathbf{cp}_k^i) } {\tilde{W}_t^l g_{k+1,j}^l},
\end{equation*}
where $\tilde{W}_o^l$, $\tilde{W}_s^l$, and $\tilde{W}_t^l$ are the corresponding weight matrices.
Finally, the residual structure combines the updated feature $\hat{g}_{k,i}^l$ and the original feature $f_{k,i}^l$ to generate the input feature for the next layer, together with a ReLU activation $f_{k,i}^{l+1} = ReLU(\hat{g}_{k,i}^l + f_{k,i}^l)$.

After $L$ layers ($l=0, \dots, L-1$), $f_{k,i}^L$ is fed into a single fully connected layer to predict the location shift $(\Delta x_k^i, \Delta y_k^i)$ for $\mathbf{cp}_k^i$. Then its coordinates can be updated as $\mathbf{\tilde{cp}}_k^i = [x_k^i + \Delta x_k^i, y_k^i + \Delta y_k^i]^T$. With the new control point location, we can get the new features $\tilde{f}_{k,i}^{l+1}$ and have it go through the whole VGCN module again to get another location shift. Such shifting of control points can be done several times to nudge the graph nodes to move to the actual region boundaries.

As shown in Figure \ref{fig:graph_conv_block}, the convolution can be viewed as spatio-temporal information integration and propagation. When several blocks are applied, each node can get information from a significantly large area of the video.


We use the Normalized Bi-directional Chamfer Distance (NBCD) for the loss to get supervision from the ground-truth boundaries. It directly measures the accuracy on keypoints and corresponds to the NBCD metric in our performance evaluation.


\section{Experiment}
\label{sec:experiment}

\subsection{Experimental settings}

\textbf{Datasets}. \textbf{Synthia} \cite{Bengar_2019_ICCV} is a recently released synthetic driving dataset that has ground truth object boundaries for every frame. This dataset contains 178 training video sequences captured at 25fps, with lengths ranging from 15 seconds to 30 seconds. We consider dynamically moving objects relevant to driving scenarios: person, car, truck, bus, and bicycle. \textbf{KITTI MOTS} \cite{2019_MOTS_CVPR} is a newly built real dataset for Multi-Object Tracking and Segmentation (MOTS). It contains 21 training sequences (12 for training, 9 for validation), and 4 testing sequences are reserved for MOTS Challenge. The dataset only has two object categories: pedestrian and car, with 99 pedestrians and 431 cars for training, 68 pedestrians and 151 cars for validation. We use the 12 training sequences for training and the validate set for testing. Synthia is about 5 times larger in terms of frame/sub-sequence numbers.     



\noindent \textbf{Evaluation Metrics}. Besides the widely used mIoU and F1-score (1px) measure \cite{Curve-GCN-CVPR19} that measures mask and boundary matching accuracy, respectively, we also use the Normalized Bi-directional Chamfer Distance (NBCD) to directly measure the keypoint matching accuracy, so that the performance can be checked from different perspectives.

\noindent \textbf{Implementation Details}
Though VGCN can handle sub-sequences of various lengths, we use equal lengths by fixing $K$ (default: 10) to ease model transfer and comparisons across-datasets and ablation study on the factor of sparsity. The 
sequence numbers of train, validation and test split for our experiments on both are 11k:0.9k:1.3k and 1.5k:0.3k:0.4k (here `k' denotes the unit of a thousand), with train and validation from the original training sets and the test split from the original test sets.



\subsection{Experimental Results}

\noindent \textbf{Models for Comparison}. Instead of video-wise joint boundary inference, one may simply apply a frame-wise model (here we choose Curve-GCN \cite{Curve-GCN-CVPR19} as it is the state-of-the-art and also the most relevant model) to each video frame with either the provided bounding box (in case of keyframes) or some interpolated bounding box (for a intermediate frame). We refer to this model as Spatial Graph Convolutional Network (SGCN), as it is also based on GCN and only does the graph convolutions spatially. Despite its simplicity, SGCN has a natural limitation of omitting the temporal relationships among successive video frames. To overcome it, one may also think about simply smoothing the results of SGCN on successive video frames using a B-spline function, so that the overall model can be made indirectly video-wise. Such a simple solution is named `SGCN-smoothed'. However, we believe that a direct modeling of temporal relationships in the model like the proposed VGCN is necessary and superior. To better show the performance difference of direct spatio-temporal modeling and indirect result smoothing, we also test a simplified version of VGCN named `VGCN-basic', by only keeping the minimal temporal connection (i.e., the decomposable local connection as shown in Figure \ref{fig:graph_structure}) and excluding the motion features. Note that for a fair comparison, all the compared models are trained with the same data which only have ground-truths on the sparse key frames. 

\noindent \textbf{Effectiveness of VGCN}. 
As shown in Table \ref{table:res_Synthia} and Table \ref{table:res_Synthia_more}, VGCN significantly outperforms SGCN under all metrics on the Synthia dataset. Interestingly, the ignorable result differences between SGCN-smoothed and SGCN also indicate that simple temporal smoothing is not effective. We found that the training set of KITTI MOTS is too small to support our model, so we choose to do the generalization experiments on it instead as we detail below.



\begin{table}[t!]
\footnotesize
\setlength\tabcolsep{2pt} 
\caption{Results on Synthia, measured by mean Intersection over Union (mIoU). Since different object categories have different amounts of samples, the number of frames in testing for each category is shown under its name.}
\label{table:res_Synthia}
\begin{center}
\begin{tabular}{ cccccccc } 
\toprule
\multicolumn{2}{c}{Model}  & Person & Car & Truck & Bus  & Bicycle & \multirow{2}{*}{Average} \\
\cmidrule{1-2} 
 Name & Property & \footnotesize{(5.96K)} & \footnotesize{(10.50K)} & \footnotesize{(0.21K)} & \footnotesize{(0.18K)}  & \footnotesize{(0.35K)} &  \\ 
\midrule
SGCN & \footnotesize{Frame-wise, based on \cite{Curve-GCN-CVPR19}}    & 68.79 & 78.17 & 78.09 & 67.93 & 32.95 & 65.19 \\ 
SGCN-smoothed & Video-wise \footnotesize{(indirect)} & 68.27 & 78.10 & 78.46 & 68.76 & 33.26 & 65.37 \\ 
  VGCN-basic  & Video-wise             & 71.75 & 79.09 & 79.91 & 65.74 & 35.53 & 66.41\\
  VGCN  & Video-wise\footnotesize{, proposed}        & \textbf{72.68} & \textbf{80.29} & \textbf{80.48} & \textbf{69.49} & \textbf{36.09} & \textbf{67.80} \\
\bottomrule
\end{tabular}
\end{center}
\end{table}

\begin{table}[t!]
\footnotesize
\caption{Results on Synthia in Normalized Bi-directional Chamfer Distance (NBCD) and F1-score (F at 2px), shown in the ``NBCD/F1-score(2px)'' format for each entry.}
\label{table:res_Synthia_more}
\setlength\tabcolsep{3pt} 
\begin{center}
\begin{tabular}{ ccccccc } 
\toprule
\multirow{2}{*}{Model}  & Person & Car & Truck & Bus  & Bicycle & \multirow{2}{*}{Average} \\
 & \footnotesize{(5.96K)} & \footnotesize{(10.50K)} & \footnotesize{(0.21K)} & \footnotesize{(0.18K)}  & \footnotesize{(0.35K)} &  \\ 
\midrule
SGCN &  3.4/86.05 & 7.3/76.86 & 5.3/80.06 & 6.1/68.43 & 11.0/50.51 & 6.6/72.39 \\ 
SGCN-smoothed &  3.4/85.50 & 7.4/76.55 & 5.2/80.07 & 6.1/69.33 & 10.7/50.03 & 6.5/72.30 \\ 
VGCN-basic & 2.9/89.89 & 7.1/80.08 & 4.9/85.11 & 5.076.13 & 10.0/56.01 & 6.0/77.44\\
VGCN   & \textbf{2.8/90.35} & \textbf{6.9/83.20} & \textbf{4.5/86.83} & \textbf{4.8/76.91} & \textbf{10.0/56.98} & \textbf{5.8/78.85} \\
\bottomrule
\end{tabular}
\end{center}
\end{table}

\begin{table}[t!]
\footnotesize
\setlength{\tabcolsep}{5pt}
\caption{Results on KITTI-MOTS. Upper part: directly applying the models trained on Synthia; lower part: fine-tuned models (pre-trained on Synthia). `Ped' stands for `Pedestrian'. }
\label{table:res_MOTS}
\begin{center}
\begin{tabular}{ cccccccccc } 
\toprule
& & \multicolumn{2}{c}{mIoU}& &\multicolumn{2}{c}{NBCD} & &\multicolumn{2}{c}{F1-score (2px)} \\
 \cmidrule{3-4} \cmidrule{6-7} \cmidrule{9-10}
 Model & & Ped & Car & & Ped & Car &  & Ped & Car \\ 
 \midrule
SGCN & & 57.1  &58.2 & &10.6 & 19.0 & & 51.5 & 52.0 \\ 
SGCN-smoothed & & 55.4  &58.1 & &10.7 & 19.0 & & 47.6 & 49.6 \\ 
VGCN-basic & & \textbf{63.3}  &70.3 & &\textbf{8.0} & 12.1 & & 60.6 & 63.4 \\ 
VGCN-basic + Full-connection & & 62.9  &\textbf{70.5} & &8.2 & \textbf{11.5} & & \textbf{62.6} & \textbf{64.0} \\ 
VGCN & & 61.7&67.2 & &9.0 & 12.0 & & 59.3 & 60.4 \\ 
 \midrule
SGCN & & 57.1  &66.1 & &9.8 & 13.7 & & 51.3 & 57.8 \\ 
SGCN-smoothed & & 55.7  &65.3 & &9.9 & 13.8 & & 48.1 & 54.9 \\ 
VGCN-basic & & 63.1  &70.9 & &8.9 & 11.8 & & 59.8 & 63.5 \\ 
VGCN-basic + Full-connection & & \textbf{65.2}  &\textbf{72.6} & &\textbf{7.8} & \textbf{11.6} & & \textbf{65.7} & \textbf{68.9} \\ 
VGCN & & 65.1  &71.8 & &8.0 & 12.0 & & 65.6 & 66.2 \\ 
 \bottomrule
\end{tabular}
\end{center}
\end{table}

\begin{figure}[t]
\begin{center}
\includegraphics[width=1.0\linewidth]{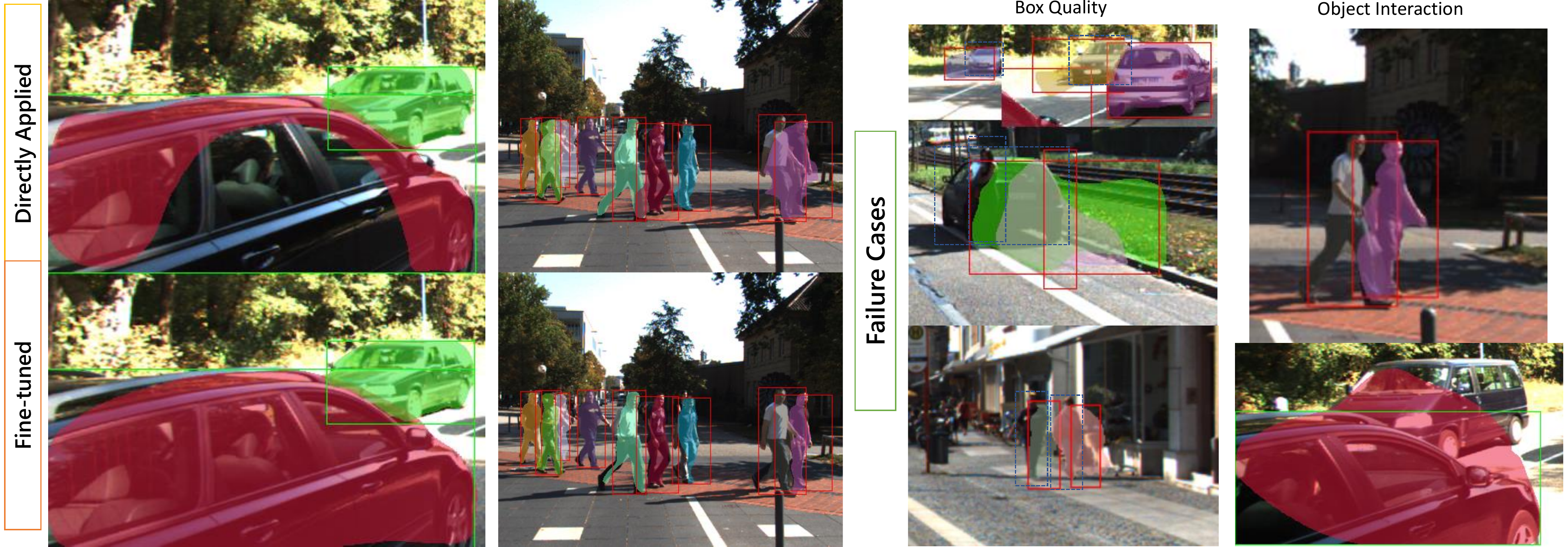}
\end{center}
\caption{Comparison of the generalizability of VGCN, and two representative failure cases.
}
\label{fig:generalization}
\end{figure}

\noindent \textbf{Generalizability}. Since the GCN-based models have no assumption on the data, they can be applied to arbitrary video data. We conduct two types of experiments to validate the generalizability of VGCN, in comparison with its competitors. One is about directly applying models trained on Synthia to the test set of KITTI MOTS. The results are shown in the upper part of Table \ref{table:res_MOTS}. The other is fine-tuning the pre-trained model (trained on Synthia) on the small training set of KITTI MOTS and then testing the fine-tuned models on its test set. The results for this are shown in the lower part of Table \ref{table:res_MOTS}. We fixed the feature extraction part, the FC layers of VGCN and its graph convolution layers, and just tuned the network for feature aggregation. VGCN models all outperform SGCN-based ones by large margins. We can see that motion features hurt VGCN in both direct application and fune-tuning, but full-connection significantly helps. Fine-tuning doesn't benefit VGCN-basic, but it significantly enhances VGCN when full-connection is adopted. These findings indicate that the major difference between synthetic data and real data is probably on the motion patterns, instead of adapting motion features which may be hard, fine-tuning the full temporal connections is more effective. Examples on how fine-tuning benefits VGCN are shown in Figure \ref{fig:generalization}, together with two representative
failure cases: undesirable region due to badly interpolated boxes and extra part caused by object interaction.

\noindent \textbf{Running Time}. Inference on a 13-frame sequence takes about 1.1s on a NVIDIA RTX2080Ti GPU. Note that interactive correction can be much faster, as features can be pre-computed and result refinement is much faster than inference-from-scratch as shown in \cite{Curve-GCN-CVPR19}. For this, we believe the computational speed of our method would be sufficient for real usage. 


\subsection{Ablation Studies}
\label{subsec:ablation}

\begin{figure}[t]
\begin{center}
  \includegraphics[width=1.0\linewidth]{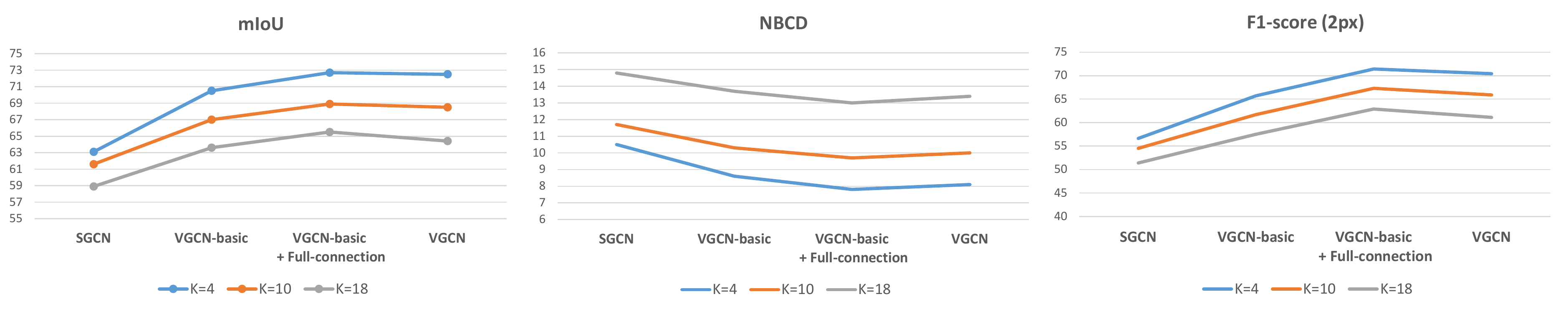}
\end{center}
\caption{Performance changes w.r.t. annotation sparsity (keyframe interval $K$).}
\label{fig:sparsity}
\end{figure}

\begin{table}[t!]
\footnotesize
\caption{Ablation studies on VGCN components using Synthia dataset, measured by mIoU.}
\label{table:ablation_synthia}
\setlength\tabcolsep{3pt} 
\begin{center}
\begin{tabular}{ ccccccc } 
\toprule
\multirow{2}{*}{Model}  & Person & Car & Truck & Bus  & Bicycle & \multirow{2}{*}{Average} \\
 & \footnotesize{(5.96K)} & \footnotesize{(10.50K)} & \footnotesize{(0.21K)} & \footnotesize{(0.18K)}  & \footnotesize{(0.35K)} &  \\ 
\midrule
VGCN-basic \footnotesize{(i.e., w/o both)} & 71.75 & 79.09 & 79.91 & 65.74 & 35.53 & 66.41\\
VGCN-basic + Motion-features &  70.40 & 79.79 & 81.20 & 67.87 & 35.34 & 66.92 \\
VGCN-basic + Full-connection  &  \textbf{73.42} & \textbf{80.30} & \textbf{81.74} & 63.92 & \textbf{36.21} & 67.12  \\ 
VGCN \footnotesize{(i.e., w/ both)} &  72.68 & 80.29 & 80.48 & \textbf{69.49} & 36.09 & \textbf{67.80} \\ 
\bottomrule
\end{tabular}
\end{center}
\end{table}


\noindent \textbf{VGCN Components}.
As shown in Table \ref{table:ablation_synthia}, both the motion features and the full-connection can help improve the performance. Though full-connection contributes more than motion features, they are complementary and the VGCN model with both performs the best.

\noindent \textbf{Sparsity of annotation}. Figure \ref{fig:sparsity} shows performance variation of the compared models on KITTI-MOTS, with different $K$ values. The results are averaged over `Ped' and `car'. All the models perform worse when $K$ is increased. VGCN models always outperform SGCN.

\noindent \textbf{Impact of box interpolation}. To investigate the influence of box quality, we use GT-bbox in testing and find the superiority of VGCN to SGCN is even greater: 68.67\% (0.87\% up) vs.65.21\%(0.02\% up) in mIoU on Synthia and 71.1\% (6.0\% up) for `Ped', 79.2\% (7.4\% up) for `Car' vs. 58.4\% (1.3\% up) and 70.7\% (4.6\% up) respectively in mIoU on KITTI MOTS.

\section{Conclusions}

This paper presents a novel tool for video region annotation that can generate dense per-frame region boundaries with only bounding boxes on sparse keyframes provided by the annotators. We believe our method opens a new avenue of research for significantly extending video supervision for general deep vision applications. An important future work is to extend the method to allow interactive correction in a human-in-the-loop annotation scheme.

\section*{Acknowledgement}
This work was in part supported by JSPS KAKENHI 17K20143 and SenseTime Japan.

\bibliography{egbib}
\end{document}